\documentclass{article}
\usepackage[preprint]{icml2026}

\makeatletter
\renewcommand{\printAffiliationsAndNotice}[1]{\global\icml@noticeprintedtrue}
\makeatother

\usepackage{fontspec}
\ifdefined\XeTeXgenerateactualtext
\fi

\usepackage{polyglossia}
\setmainlanguage{english}
\setotherlanguage{hindi}

\newfontfamily\hindifont[
  Path = fonts/,
  Extension = .ttf,
  UprightFont = *-Regular,
  BoldFont = *-Bold,
  Script = Devanagari,
  Scale = 0.94
]{NotoSerifDevanagari}

\newfontfamily\devanagarifont[
  Path = fonts/,
  Extension = .ttf,
  UprightFont = *-Regular,
  BoldFont = *-Bold,
  Script = Devanagari,
  Scale = 0.94
]{NotoSerifDevanagari}

\providecommand{\texthindi}[1]{{\hindifont #1}}

\newcommand{\RETURN}{\STATE \textbf{return}\ }

\usepackage{microtype}
\usepackage{graphicx}
\usepackage{booktabs}
\usepackage{multirow}
\usepackage{makecell}
\usepackage{array}
\usepackage{amsmath,amssymb}
\usepackage{float}
\usepackage{algorithm}
\usepackage{algorithmic}
\usepackage{xcolor}
\usepackage[hidelinks]{hyperref}
\hypersetup{
  pdfauthor={Samyak Savi, Chavi Gupta, Shreyas Gantayet, Tanay Sodha, Dhruv Kumar}
}
\usepackage{url}

\usepackage{tabularx}
\usepackage{stfloats}
\usepackage{tikz}
\usetikzlibrary{arrows.meta,positioning}

\definecolor{goodgreen}{RGB}{44,160,44}
\definecolor{badred}{RGB}{214,39,40}
\definecolor{lightgreen}{RGB}{200,240,210}
\definecolor{lightred}{RGB}{250,210,207}

\icmltitlerunning{Cultural Fidelity in Hindi MT}

\begin{document}

\twocolumn[
\icmltitle{Cultural Fidelity in English-to-Hindi Translation:\\
A Preservation--Fluency Frontier for Gender Recoverability}
\icmltitlerunning{Cultural Fidelity in Hindi MT}
\begin{center}
{\bf Samyak Savi\textsuperscript{1}, Chavi Gupta\textsuperscript{1}, Shreyas Gantayet\textsuperscript{1}, Tanay Sodha\textsuperscript{1}, Dhruv Kumar\textsuperscript{1}}\\
\textsuperscript{1}BITS Pilani
\end{center}
\vskip 0.3in
]

\printAffiliationsAndNotice{}

\begin{abstract}
Generative translation systems are cultural technologies because they decide
how socially meaningful cues are rendered within culturally specific
grammatical systems. We study one concrete notion of successful cultural
translation: when an English source explicitly encodes gender, an
English-to-Hindi translation should preserve the recoverability of that cue
unless the source itself is ambiguous. We evaluate this criterion on a
37,345-instance benchmark spanning twelve categories and show that five
systems frequently erase gender through ergative and honorific constructions.
We then introduce two mechanism-aware inference-time interventions. The first,
the Source-Aware Reranker (SAR), prefers candidates that avoid
gender-neutralizing syntax. The second, the Phenomenon-Aware Reranker (PAR),
preserves gender through targeted lexical marking even when ergative syntax
remains. Across GPT-4o-mini and Sarvam, PAR improves target-subset accuracy
from 11.07\% to 54.47\% and from 15.99\% to 49.66\%, respectively. Human
evaluation shows that PAR increases gender preservation from 10.3\% to
81.3\%, but reduces mean fluency from 4.36 to 3.37. These findings place the two
interventions on a preservation and fluency frontier rather than supporting a
single dominant solution, and show how culturally situated generation can
require explicit tradeoffs among fidelity, fluency, and stylistic naturalness.

\end{abstract}

\section{Introduction}
\label{sec:intro}

Translation models do not merely transfer information across languages. They
also decide how socially meaningful cues are expressed in culturally specific
target grammars. This makes translation a useful case study for cultural AI.
A positive notion of success cannot stop at avoiding harmful bias. It must
also ask how a system should faithfully render identity-relevant information
when the target language encodes that information differently from the source
language.

Unlike prior gender-bias evaluations that primarily ask whether systems
reproduce stereotypes, we study a narrower but culturally grounded success
condition. When the source speaker explicitly encodes gender, the translation
should preserve the recoverability of that cue unless the source itself is
ambiguous. This reframes gender handling as an interpretive design problem
rather than only a harm-detection problem. The translation system must decide
how a socially meaningful source cue should survive in a target grammar that
may otherwise erase it.

We study this question for English-to-Hindi translation. English often leaves
grammatical gender underspecified, while Hindi expresses gender through
agreement on verbs, adjectives, and pronouns. This difference becomes
particularly consequential in perfective constructions, where Hindi exhibits
ergative alignment. A fluent translation of ``She completed the project'' can
appear as \texthindi{उसने प्रोजेक्ट पूरा किया}, where agreement shifts to the
masculine object and the feminine subject is no longer recoverable. The
output is grammatical, but explicit source-side identity information has been
erased.

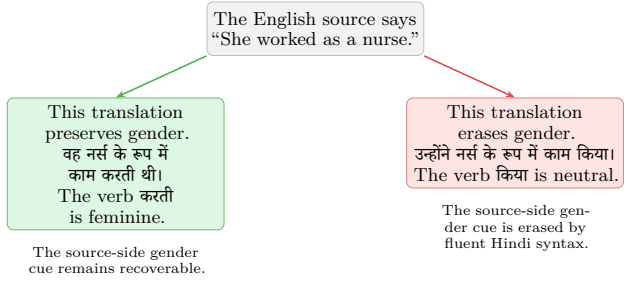
\begin{figure}[t]
\centering
\resizebox{\columnwidth}{!}{%
\begin{tikzpicture}[
  font=\small,
  node distance=0.5cm,
  box/.style={rectangle, rounded corners=3pt, draw=gray!50,
              align=center, fill=white, minimum width=3.6cm,
              minimum height=0.9cm, text width=3.4cm},
  greenbox/.style={box, fill=lightgreen!60, draw=goodgreen!70},
  redbox/.style={box, fill=lightred!60, draw=badred!60},
  graybox/.style={box, fill=gray!10},
  arrow/.style={-{Stealth[length=4.5pt]}, thick},
  label/.style={font=\scriptsize, align=center}
]

\node[graybox] (src) {The English source says\\``She worked as a nurse.''};

\node[greenbox, below left=0.7cm and -0.3cm of src] (good)
  {This translation preserves gender.\\\texthindi{वह नर्स\ के रूप में काम करती थी।}\\
  The verb \texthindi{करती} is feminine.};

\node[redbox, below right=0.7cm and -0.3cm of src] (bad)
  {This translation erases gender.\\\texthindi{उन्होंने\ नर्स\ के रूप में काम किया।}\\
  The verb \texthindi{किया} is neutral.};

\draw[arrow, goodgreen!80] (src.south west) -- (good.north);
\draw[arrow, badred!80] (src.south east) -- (bad.north);

\node[label, below=0.15cm of good, text width=3.4cm]
  {The source-side gender cue remains recoverable.};
\node[label, below=0.15cm of bad, text width=3.4cm]
  {The source-side gender cue is erased by fluent Hindi syntax.};
\end{tikzpicture}
}
\caption{Hindi permits more than one fluent translation for an explicitly
gendered English sentence. The gender-preserving path marks the subject
through verbal agreement, while the ergative or honorific path produces a
grammatical sentence in which the subject's gender is no longer recoverable.}
\label{fig:ergative_mechanism}
\end{figure}

Figure~\ref{fig:ergative_mechanism} illustrates the mechanism. Both Hindi
translations are fluent, but only one makes the feminine subject recoverable.
The second translation is not an ungrammatical error. It is precisely the
kind of grammatical output that creates a cultural-fidelity failure.

Prior work documents gender bias in machine translation, mostly for European
target languages or broader multilingual settings
\citep{stanovsky2019evaluating,savoldi2021gender,currey2022mtgeneval,robinson2024mittens}.
Hindi-specific work also exists
\citep{ramesh2021evaluating,singh2023gender,hada2024akal}. Unlike
prior mitigation work, which often targets wrong gender inflection on a
specific word, we target grammatical erasure through ergative and honorific
constructions, and we evaluate two distinct mitigation strategies for this
mechanism. A system can avoid the neutralizing construction, or it can bypass
the construction by adding a minimal lexical gender marker that survives it.
We evaluate both strategies automatically and through human judgment.

This paper makes three contributions. First, we define a concrete positive
cultural objective for translation: source-explicit identity cues should
remain recoverable in the target language unless the source is ambiguous.
Second, we identify a Hindi-specific mechanism of cultural-fidelity failure,
where fluent ergative and honorific constructions erase source-side gender
without producing an ungrammatical translation. Third, we show that
mitigation is not a single optimization problem: an avoidance-style reranker
and a bypass-style lexicalized reranker occupy different points on a
preservation--fluency frontier, validated by both automatic evaluation and
blinded Hindi-fluent human judgment.

\section{Related Work}
\label{sec:related}

\paragraph{Gender bias evaluation in machine translation.}
WinoMT introduced a controlled challenge set and automatic evaluation
protocol for measuring gender bias in machine translation across
grammatical-gender target languages \citep{stanovsky2019evaluating}.
MT-GenEval later expanded gender evaluation with counterfactual and
contextual data, including multi-sentence settings that require agreement
beyond a single local phrase \citep{currey2022mtgeneval}. MiTTenS further
broadened evaluation to a larger multilingual setting and studied
misgendering harms across dedicated translation systems and foundation models
\citep{robinson2024mittens}. These benchmarks establish the importance of
gender-sensitive MT evaluation, but they do not focus on the Hindi-specific
mechanism in which fluent ergative or honorific syntax can erase source-side
gender.

\paragraph{Hindi and Indic gender bias.}
Prior work on Hindi and Indic language technology shows that grammatical
gender creates distinctive measurement challenges. Studies of Hindi-English
and Hindi-related translation observe that surface forms can change with
gender and that evaluation must account for morphology rather than only
lexical choice \citep{ramesh2021evaluating,singh2023gender}. Broader studies
of Hindi language technology also emphasize community-centric and Global
South perspectives on gender bias \citep{hada2024akal}. Kirtane and
Anand study mitigation for gender stereotypes in Hindi and Marathi
\citep{kirtane2022mitigating}. Our work builds on this line, but studies the
opposite direction, English-to-Hindi, where the target language must decide
whether and how to realize gender that was explicit in the source.

\paragraph{Mitigation through annotations, decoding, and prompting.}
Several prior studies move beyond diagnosis. Target-gender annotations have
been used to train MT systems that more reliably express the intended gender
\citep{stafanovics2020mitigating}. Black-box and inference-time interventions have also been proposed
for gender control, including context injection \citep{moryossef2019filling},
source-side tagging \citep{vanmassenhove2018getting}, and n-best gender-aware
reranking \citep{saunders2020neural,saunders2022first}. Prompting-based work
finds that LLM prompts can reduce some gender-bias failures, but that their
effectiveness depends on the failure mode \citep{attanasio2023tale,sant2024power}.
Our inference-time approach is closest to gender-aware reranking, but differs
in targeting English-to-Hindi ergative and honorific neutralization. The
mitigation problem is not only to select the correct gendered inflection, but
also to decide whether to avoid the neutralizing construction or to bypass it
through explicit lexical marking.

\paragraph{Cultural translation and cultural fidelity.}
Recent work on cultural machine translation argues that translation quality
must include pragmatic and cultural appropriateness, not only literal
semantic adequacy \citep{yao2024benchmarking}. Related work on cultural
adaptation for language models studies how systems can adapt text to
context-specific cultural expectations \citep{singh2024translating}. Our
study contributes to this direction by defining a culturally grounded success
condition for English-to-Hindi translation: source-side gender cues should
remain recoverable when they are explicitly expressed.

\section{Cultural Framing and Benchmark Setting}
\label{sec:benchmark}

This work articulates a positive notion of success for cultural AI. In our
setting, that notion is narrow by design. We
do not ask the model to infer hidden identity information. We only ask it to
preserve a cue the source has already chosen to express. This objective
extends rather than replaces speaker agency. It is therefore compatible with
a constructive cultural framing rather than a purely harm-mitigation framing.

This setting is an instance of interpretive technology. The reranker is not
merely optimizing an automatic accuracy metric. It operationalizes an
interpretive judgment about when a source-side identity cue should remain
recoverable in Hindi. The intervention therefore makes a cultural design
choice explicit: a translation may remain fluent while failing to preserve an
identity-relevant cue that the source deliberately expressed. We treat this
as a constructive cultural objective, rather than only as a bias-detection
problem.

Our benchmark contains 37,345 English-to-Hindi examples across twelve
linguistically motivated categories, including explicit gender, late binding,
Winograd-style coreference, profession stereotypes, and name-based cues. The
benchmark uses 45 professions across four stereotype classes, 30 Indian
names, and 50 manually curated cross-stereotype profession pairs following
the WinoBias methodology \citep{zhao2018wino}. Ground truth labels are
assigned deterministically from source-side content.

\begin{table}[!b]
\caption{Baseline diagnostic summary for unreranked MT outputs. Overall
accuracy is measured on the full 37,345-instance benchmark. Ergative rate is
the proportion of outputs containing the ergative marker \texthindi{ने}.}
\label{tab:diagnostic_summary}
\centering
\footnotesize
\setlength{\tabcolsep}{5pt}
\begin{tabular}{lrr}
\toprule
Baseline MT system & Overall Acc. & Ergative Rate \\
\midrule
Helsinki & 47.96 & 63.2 \\
NLLB-200 & 57.23 & 76.2 \\
IndicTrans2 & 58.73 & 74.7 \\
Sarvam & 58.43 & 74.9 \\
GPT-4o-mini & 57.18 & 76.1 \\
\bottomrule
\end{tabular}
\end{table}

We use the benchmark dataset described above and a V2 grammar-aware
Hindi classifier. The classifier combines explicit gender-word detection,
bilingual name lookup, ergative-marker detection, singular verb morphology,
honorific plural detection, and an LLM fallback. In validation on 100 doubly
annotated examples, this classifier achieved 83.8\% agreement with human
labels overall and 93.9\% agreement when predicting neutral. The automatic
classifier is used as a large-scale diagnostic instrument rather than as a
complete substitute for human judgment. This distinction is especially
important for PAR, because PAR can insert lexical gender markers that the
automatic classifier is also designed to detect. We therefore treat automatic
results as directional large-scale evidence and use blinded human evaluation as
the main validity check for the preservation and fluency tradeoff. The
classifier validation supports the reliability of the neutralization signal,
but the human study remains the primary evidence for the preservation and
fluency frontier.

We focus mitigation on a target subset of 15,750 examples formed by the
union of explicit\_gender, late\_binding, and winograd\_coref rows whose
English source contains recoverable male or female cues. This subset isolates
the cases where the source explicitly commits to a gender and where
preservation, rather than stereotype avoidance alone, is the central cultural
objective. The remaining categories involve implicit, ambiguous, or
stereotype-based cues for which binary source-cue preservation is not the
right mitigation objective.
The evaluated baselines include Helsinki OPUS-MT \citep{tiedemann2020opusmt},
NLLB-200 \citep{nllbteam2022}, IndicTrans2 \citep{gala2024indictrans2},
Sarvam \citep{sarvam2024}, and GPT-4o-mini \citep{openai2024gpt4omini}.

\section{Mechanism-Aware Mitigation Methods}
\label{sec:methods}

We study two inference-time interventions: the Source-Aware Reranker (SAR)
and the Phenomenon-Aware Reranker (PAR). Both methods generate multiple Hindi
candidates and rerank them without changing model weights, but they target the
Hindi gender-erasure mechanism in different ways. For both base
systems, candidate generation in the reranking stage uses GPT-4o-mini. When
the base system is GPT-4o-mini, the reranker selects among GPT-4o-mini
candidates. When the base system is Sarvam, the reranker selects from the
original Sarvam translation plus GPT-4o-mini candidates. We therefore
interpret the Sarvam experiments as evaluating a GPT-assisted reranking layer
applied to Sarvam outputs, not as a pure Sarvam-only generation method. This
design is intentional because our goal is to test whether a mechanism-aware
intervention can repair gender erasure in existing translations. It also
means that cross-system comparisons should be read as parallel case studies
rather than as a strict comparison of independently generated candidate pools.
We call the first method source-aware because it scores candidates against the
source-side gender cue, and the second phenomenon-aware because it changes
prompts and lexicalization strategy according to the specific erasure
phenomenon. Intuitively, SAR asks whether we can choose a fluent candidate
that avoids the erasing construction. PAR asks whether, when such
constructions remain natural, the source cue can still be made recoverable
through minimal lexical marking.

\paragraph{SAR avoids neutralizing constructions.}
SAR generates $k=5$ Hindi candidates from GPT-4o-mini and reranks them using
a quality score, a source-side gender-preservation score, and an optional
ergative penalty. SAR mainly works by avoiding the
ergative neutralization mechanism. It reduces target-subset ergative
usage from 84.5\% to 75.5\% in the Sarvam repair setting and from 87.4\% to
75.9\% in the GPT-4o-mini setting.

\paragraph{PAR with lexicalization bypasses neutralization.}
PAR is stronger and more targeted. It detects whether the English source is an
explicit-gender, late-binding, or Winograd-style example using only source-side
cues. PAR also generates $k=5$ GPT-4o-mini candidates, but the prompt
and scoring rule are selected according to the detected phenomenon. It then
issues phenomenon-specific reranking prompts and allows minimal lexical gender
marking, such as \texthindi{पुरुष} or \texthindi{महिला}, when grammatical
agreement alone would otherwise erase the source cue. PAR mainly works by
bypassing ergative neutralization. It keeps ergative usage near the
corresponding Baseline, but it makes the gender recoverable through lexical
marking that survives the construction.

\paragraph{The factorial ablation separates lexicalization from reranking.}
To separate the contribution of lexicalization from the contribution of
phenomenon-aware reranking, we ran a $2 \times 2$ factorial experiment on the
target subset for GPT-4o-mini. Lexicalization alone mainly helps
explicit\_gender. Phenomenon-aware reranking alone mainly helps
late\_binding and winograd\_coref. The full combination is strongest overall.
This pattern shows that PAR is not a single undifferentiated prompt effect.

\paragraph{The human evaluation measures preservation, fluency, and preference.}
We evaluate PAR on 150 stratified target-subset examples,
with 50 examples from each target category. Two Hindi-fluent annotators
compare blinded Baseline and reranked outputs. Annotators see the English
source sentence because preservation is defined relative to source-side
gender evidence, but the two Hindi outputs are shuffled and anonymized. They
label whether each output preserves source gender, rate Hindi fluency on a
1 to 5 scale using grammaticality and naturalness rather than preservation,
and choose a preferred output. Ties are allowed and excluded from the
non-tie preference rate. For comparison, we also report SAR human-evaluation
results collected with the same protocol on 150 examples.

\section{Results}
\label{sec:results}

\subsection{Prompting is Not Enough}

Before turning to reranking, we summarize the negative result that motivates
it. On Sarvam, Level 3 prompting, the strongest prompt-only setting
with an explicit gender-preservation instruction, improves overall benchmark
accuracy from 58.43\% to 60.42\%, but reduces target-subset accuracy from
15.99\% to 14.63\%. The same pattern appears in GPT-4o-mini target-subset experiments,
where lexicalized prompting alone helps explicit-gender rows but leaves
multi-clause pronoun resolution largely unresolved. Prompting therefore
redistributes errors, but it does not repair the underlying mechanism.

\subsection{Repairing Outputs from Two Base Systems}

\begin{table*}[t]
\caption{Main mitigation results. Baseline denotes unreranked MT output. SAR
is the Source-Aware Reranker, and PAR is the Phenomenon-Aware Reranker.
Target accuracy is evaluated on the 15,750 target-subset rows; full accuracy
is evaluated on all 37,345 rows. Ergative denotes the percentage of
target-subset outputs containing the ergative marker \texthindi{ने}.}
\label{tab:main_results}
\centering
\small
\setlength{\tabcolsep}{4.5pt}
\begin{tabular}{lrrrrrr}
\toprule
System / intervention & Exp. gender & Late binding & Winograd & Target Acc. & Full Acc. & Ergative \\
\midrule
Sarvam Baseline & 26.7 & 7.8 & 5.7 & 15.99 & 58.43 & 84.5 \\
Sarvam Prompted L3 & 26.0 & 5.6 & 3.9 & 14.63 & 60.42 & 83.3 \\
Sarvam + SAR & 61.0 & 9.7 & 5.3 & 32.47 & 65.38 & 75.5 \\
Sarvam + PAR & 59.6 & 27.2 & 45.7 & 49.66 & 72.68 & 83.4 \\
\addlinespace
GPT Baseline & 21.1 & 7.2 & 0.0 & 11.07 & 57.18 & 87.4 \\
GPT + SAR & 59.4 & 9.2 & 1.5 & 30.17 & 65.04 & 75.9 \\
GPT + PAR & 64.4 & 49.5 & 44.0 & 54.47 & 74.48 & 84.0 \\
\bottomrule
\end{tabular}
\end{table*}

PAR improves both evaluated repair settings, as shown in
Table~\ref{tab:main_results}. Relative to Baseline MT output, target-subset accuracy
rises by 33.67 points on Sarvam and 43.40 points on GPT-4o-mini.
Full-benchmark accuracy also improves strongly, reaching 72.68\% on Sarvam
and 74.48\% on GPT-4o-mini. Paired McNemar tests confirm that these paired
improvements are statistically significant in both systems. On Sarvam, for example, PAR beats
the corresponding Baseline output on 5,356 target-subset rows while losing on only 53.

The per-category pattern is equally important. Lexicalized prompting alone
helps explicit-gender cases but leaves multi-clause pronoun resolution
largely unsolved. PAR is what moves late\_binding
and winograd\_coref. Across base systems, winograd\_coref converges to
roughly 45\% under PAR, up from 0.0\% for the GPT Baseline and
5.7\% for the Sarvam Baseline. This is evidence that PAR is targeting the
underlying mechanism rather than one model's idiosyncrasies.
\subsection{PAR Factorial Ablation}

Table~\ref{tab:par_ablation} shows that PAR is not a single prompt effect.
Lexicalization alone improves explicit-gender cases, but barely helps
late-binding or Winograd-style examples. Phenomenon-aware prompts alone nearly
saturate late-binding accuracy in this ablation, but damage explicit-gender
examples where simpler agreement morphology was already sufficient. Only the combination
improves all three target categories simultaneously.

This interaction suggests that PAR combines two separable functions:
source-side cue localization and target-side cue preservation.
Phenomenon-aware prompting helps locate delayed or coreferential gender cues,
while lexicalization provides a robust way to keep those cues recoverable in
Hindi. A separate SAR ablation, reported in
Appendix~\ref{app:method_details}, shows that quality-only reranking remains
near Baseline and that gender-aware scoring drives SAR's gains.

\begin{table}[b]
\caption{PAR factorial ablation on the GPT target subset. Lexicalization mainly
helps explicit-gender cases, while phenomenon-aware prompts are necessary for
late-binding and Winograd-style cases. The No/No row is a candidate-pool
ablation control, not the raw GPT Baseline in Table~\ref{tab:main_results}.}
\label{tab:par_ablation}
\centering
\scriptsize
\setlength{\tabcolsep}{3pt}
\begin{tabular}{llrrrr}
\toprule
Lex. & Phen. prompts & Exp. & Late & Wino. & Target \\
\midrule
No  & No  & 40.2 & 4.8   & 2.6  & 20.8 \\
Yes & No  & 57.0 & 3.6   & 6.4  & 30.1 \\
No  & Yes & 23.7 & 100.0 & 18.5 & 32.6 \\
Yes & Yes & 64.3 & 49.5  & 44.0 & 54.5 \\
\bottomrule
\end{tabular}
\end{table}

\subsection{Human Evaluation and the Preservation-Fluency Frontier}

We conducted blinded human evaluation on 150 stratified examples, with 50
examples from each target category. Two Hindi-fluent annotators saw the
English source and two anonymized Hindi outputs in randomized order, then
judged gender-cue preservation, fluency, and overall preference.
Appendix~\ref{app:human_protocol} gives the full protocol.

\begin{figure}[t]
\centering
\begin{tikzpicture}[x=3.55cm,y=0.038cm,font=\footnotesize]
  \draw[->, thick] (-1.15,-8) -- (0.22,-8);
  \node[below=11pt] at (-0.5,-8) {Fluency change on 1--5 scale};
  \draw[->, thick] (-1.15,-8) -- (-1.15,84);
  \node[above, rotate=90, anchor=south] at (-1.31,40) {Preservation gain (pp)};
  \draw[dashed, gray!45] (0,-4) -- (0,84);
  \draw[dashed, gray!45] (-1.15,0) -- (0.22,0);
  \foreach \x/\lab in {-1.0/-1.0,-0.8/-0.8,-0.6/-0.6,-0.4/-0.4,-0.2/-0.2,0/0.0}
    \draw (\x,-8) -- (\x,-6) node[below=3pt] {\lab};
  \foreach \y in {0,20,40,60,80}
    \draw (-1.15,\y) -- (-1.13,\y) node[left=3pt] {\y};
  \draw[densely dashed, gray!60, line width=0.7pt] (-0.99,71.0) -- (-0.01,8.3);
  \filldraw[fill=gray!60, draw=black] (0,0) circle[radius=0.018];
  \node[gray!70!black, font=\footnotesize\bfseries, right=4pt] at (0,0) {Baseline};
  \draw[fill=blue!70!black, draw=black] (-0.01,8.3) rectangle ++(-0.035,3.1);
  \node[blue!70!black, font=\footnotesize\bfseries, above left=1pt] at (-0.01,11.4) {SAR};
  \draw[fill=red!75!black, draw=black] (-0.99,71.0) -- ++(-0.025,-3.2) -- ++(0.05,0) -- cycle;
  \node[red!75!black, font=\footnotesize\bfseries, above right=1pt] at (-0.99,71.0) {PAR};
\end{tikzpicture}
\caption{Preservation--fluency frontier from human evaluation. SAR provides a
small preservation gain with almost no fluency cost, while PAR provides a much
larger preservation gain at a substantial fluency cost. Negative fluency change
indicates lower mean fluency relative to Baseline.}
\label{fig:frontier}
\end{figure}
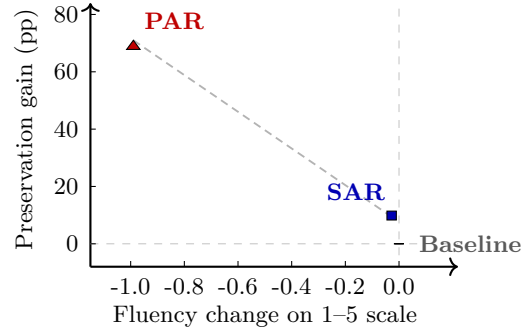

The human results support the preservation and fluency frontier framing. PAR
produces a large human-validated preservation gain, increasing preservation
from 10.3\% to 81.3\%, but it also reduces mean fluency from 4.36 to 3.37.
Bootstrap 95\% confidence intervals support the same conclusion:
preservation rises from 10.3\% [7.0, 14.0] to 81.3\% [76.7, 85.7], while
mean fluency falls from 4.36 [4.28, 4.43] to 3.37 [3.22, 3.50].
Overall preference was roughly balanced for PAR: annotators preferred the
Baseline in 42.3\% of judgments and PAR in 39.3\%, with the remaining
judgments marked as ties.

Figure~\ref{fig:frontier} visualizes the central tradeoff. SAR occupies a
conservative point (+8.3 preservation points, essentially no fluency cost),
while PAR occupies a high-preservation point (+71.0 points, -0.99 fluency).
The tradeoff is category-dependent: late-binding examples have the largest
fluency drop (-1.90), whereas explicit-gender examples show almost no fluency
loss (-0.09). This pattern explains why PAR is not a dominant solution; it is
attractive for explicit-gender cases but can preserve late-binding cues through
awkward lexicalization. Appendix~\ref{app:human_protocol} reports the category
breakdown.

\section{Discussion}
\label{sec:discussion}

These results suggest a concrete positive notion of cultural success for
translation systems. In this setting, success is not simply the avoidance of
harmful stereotypes. It is the faithful rendering of an explicitly expressed
source-side identity cue within the grammatical resources of a culturally
specific target language. The benchmark diagnosis shows that present systems
often fail at this. The mitigation results show that the failure is not
immutable.

At the same time, the human evaluation makes clear that preservation is not
the only relevant value. PAR works because it
inserts overt lexical gender markers that survive ergative syntax. This is
effective, but sometimes stylistically marked. We therefore do not present
PAR as a dominant solution. Instead, SAR and PAR trace a preservation and
fluency frontier that makes the tradeoff explicit and supports
context-dependent operating points.

This frontier also clarifies method choice. SAR is the better default when
readability and minimal stylistic disruption matter most; PAR becomes
attractive when faithful rendering of explicit identity cues is the priority.
We view this as a configuration problem rather than a single-metric
optimization problem.

\paragraph{The study has four limitations.}
First, PAR uses explicit lexical gender marking, which is a stylistic
intervention rather than a universally preferred Hindi translation strategy.
Human evaluation shows that this intervention substantially improves
preservation, but also introduces a meaningful fluency cost, especially in
late-binding constructions. Second, both reranking methods use GPT-4o-mini
candidates, even when repairing Sarvam translations. The Sarvam results should
therefore be interpreted as GPT-assisted reranking of Sarvam outputs. Third,
the benchmark is constructed and diagnostic, so naturalistic validation remains
necessary before deployment claims can be made. Fourth, the automatic evaluator
and the reranker share some morphology features. We mitigate this concern
through blinded human evaluation, but the automatic metrics should still be
interpreted as large-scale directional evidence rather than as final judgments
of translation quality.

\section{Conclusion}
\label{sec:conclusion}

We studied English-to-Hindi gender preservation as a culturally grounded
translation objective. The core failure is not only wrong gender assignment,
but also grammatical erasure through ergative and honorific constructions.
SAR improves preservation by selecting candidates that reduce
gender-neutralizing syntax, while PAR improves preservation more aggressively
through minimal lexical markers that survive ergative syntax. Automatic
results show large gains across GPT-4o-mini and Sarvam; human evaluation
shows that PAR substantially improves preservation while reducing fluency,
especially in late-binding constructions. Culturally faithful translation
should therefore be studied as a configurable preservation and fluency
frontier rather than as a single accuracy-maximization problem.

\begingroup
\small
\sloppy
\bibliography{references}

\begin{thebibliography}{23}
\providecommand{\natexlab}[1]{#1}
\providecommand{\url}[1]{\texttt{#1}}
\expandafter\ifx\csname urlstyle\endcsname\relax
  \providecommand{\doi}[1]{doi: #1}\else
  \providecommand{\doi}{doi: \begingroup \urlstyle{rm}\Url}\fi

\bibitem[Attanasio et~al.(2023)Attanasio, Plaza~del Arco, Nozza, and
  Lauscher]{attanasio2023tale}
G.~Attanasio, F.~M. Plaza~del Arco, D.~Nozza, and A.~Lauscher.
\newblock A tale of pronouns: Interpretability informs gender bias mitigation
  for fairer instruction-tuned machine translation.
\newblock In \emph{Proceedings of the 2023 Conference on Empirical Methods in
  Natural Language Processing}, pages 3996--4014, Singapore, 2023. Association
  for Computational Linguistics.
\newblock \doi{10.18653/v1/2023.emnlp-main.243}.
\newblock URL \url{https://aclanthology.org/2023.emnlp-main.243/}.

\bibitem[Currey et~al.(2022)Currey, Nadejde, Pappagari, Mayer, Lauly, Niu, Hsu,
  and Dinu]{currey2022mtgeneval}
A.~Currey, M.~Nadejde, R.~R. Pappagari, M.~Mayer, S.~Lauly, X.~Niu, B.~Hsu, and
  G.~Dinu.
\newblock {MT-GenEval}: A counterfactual and contextual dataset for evaluating
  gender accuracy in machine translation.
\newblock In \emph{Proceedings of the 2022 Conference on Empirical Methods in
  Natural Language Processing}, pages 4287--4299. Association for Computational
  Linguistics, 2022.
\newblock \doi{10.18653/v1/2022.emnlp-main.288}.

\bibitem[Gala et~al.(2024)Gala, Chitale, Raghavan, Dhore, Sureja, Doddapaneni,
  Bapna, Ramesh, Kunchukuttan, Kumar, et~al.]{gala2024indictrans2}
J.~Gala, P.~A. Chitale, A.~Raghavan, V.~Dhore, S.~Sureja, S.~Doddapaneni,
  A.~Bapna, G.~Ramesh, A.~Kunchukuttan, P.~Kumar, et~al.
\newblock {IndicTrans2}: Towards high-quality and accessible machine
  translation models for all 22 scheduled {Indian} languages.
\newblock \emph{Transactions on Machine Learning Research}, 2024.

\bibitem[Hada et~al.(2024)Hada, Husain, Gumma, Diddee, Yadavalli, Seth,
  Kulkarni, Gadiraju, Vashistha, Seshadri, and Bali]{hada2024akal}
R.~Hada, S.~Husain, V.~Gumma, H.~Diddee, A.~Yadavalli, A.~Seth, N.~Kulkarni,
  U.~Gadiraju, A.~Vashistha, V.~Seshadri, and K.~Bali.
\newblock Akal badi ya bias: An exploratory study of gender bias in {H}indi
  language technology.
\newblock In \emph{Proceedings of the 2024 ACM Conference on Fairness,
  Accountability, and Transparency}, 2024.
\newblock \doi{10.1145/3630106.3659017}.

\bibitem[Kirtane and Anand(2022)]{kirtane2022mitigating}
\begingroup\color{blue}
N.~Kirtane and T.~Anand.
\newblock \revised{Mitigating Gender Stereotypes in Hindi and Marathi}.
\newblock In \emph{\revised{Proceedings of the 4th Workshop on Gender Bias in
  Natural Language Processing}}. \revised{Association for Computational
  Linguistics}, 2022.
\endgroup

\bibitem[Moryossef et~al.(2019)Moryossef, Aharoni, and
  Goldberg]{moryossef2019filling}
\begingroup\color{blue}
A.~Moryossef, R.~Aharoni, and Y.~Goldberg.
\newblock \revised{Filling Gender and Number Gaps in Neural Machine Translation
  with Black-box Context Injection}.
\newblock In \emph{\revised{Proceedings of the First Workshop on Gender Bias in
  Natural Language Processing}}. \revised{Association for Computational
  Linguistics}, 2019.
\endgroup

\bibitem[{NLLB Team} et~al.(2022){NLLB Team}, Costa-juss{\`a}, Cross,
  {\c{C}}elebi, Elbayad, Heafield, et~al.]{nllbteam2022}
{NLLB Team}, M.~R. Costa-juss{\`a}, J.~Cross, O.~{\c{C}}elebi, M.~Elbayad,
  K.~Heafield, et~al.
\newblock No language left behind: Scaling human-centered machine translation.
\newblock Technical report, Meta AI, 2022.
\newblock arXiv:2207.04672.

\bibitem[{OpenAI}(2024)]{openai2024gpt4omini}
{OpenAI}.
\newblock {GPT-4o} mini: Advancing cost-efficient intelligence, 2024.
\newblock OpenAI product announcement. Accessed: March 2026.

\bibitem[Ramesh et~al.(2021)Ramesh, Gupta, and Singh]{ramesh2021evaluating}
K.~Ramesh, G.~Gupta, and S.~Singh.
\newblock Evaluating gender bias in {H}indi-{E}nglish machine translation.
\newblock In \emph{Proceedings of the 3rd Workshop on Gender Bias in Natural
  Language Processing}, pages 16--23. Association for Computational
  Linguistics, 2021.
\newblock \doi{10.18653/v1/2021.gebnlp-1.3}.

\bibitem[Robinson et~al.(2024)Robinson, Kudugunta, Stella, Dev, and
  Bastings]{robinson2024mittens}
K.~Robinson, S.~Kudugunta, R.~Stella, S.~Dev, and J.~Bastings.
\newblock {MiTTenS}: A dataset for evaluating gender mistranslation.
\newblock In \emph{Proceedings of the 2024 Conference on Empirical Methods in
  Natural Language Processing}, pages 4115--4124. Association for Computational
  Linguistics, 2024.

\bibitem[Sant et~al.(2024)Sant, Escolano, Mash, De~Luca~Fornaciari, and
  Melero]{sant2024power}
\begingroup\color{blue}
A.~Sant, C.~Escolano, A.~Mash, F.~De~Luca~Fornaciari, and M.~Melero.
\newblock \revised{The Power of Prompts: Evaluating and Mitigating Gender Bias
  in MT with LLMs}.
\newblock In \emph{\revised{Proceedings of the 5th Workshop on Gender Bias in
  Natural Language Processing}}, pages 94--139, Bangkok, Thailand, 2024.
  \revised{Association for Computational Linguistics}.
\newblock \doi{10.18653/v1/2024.gebnlp-1.7}.
\newblock URL \url{https://aclanthology.org/2024.gebnlp-1.7/}.
\endgroup

\bibitem[{Sarvam AI}(2024)]{sarvam2024}
{Sarvam AI}.
\newblock Sarvam translate.
\newblock \url{https://www.sarvam.ai/blogs/sarvam-translate}, 2024.
\newblock Accessed: March 2026.

\bibitem[Saunders et~al.(2020)Saunders, Sallis, and Byrne]{saunders2020neural}
D.~Saunders, R.~Sallis, and B.~Byrne.
\newblock Neural machine translation doesn't translate gender coreference right
  unless you make it.
\newblock In \emph{Proceedings of the Second Workshop on Gender Bias in Natural
  Language Processing}, pages 35--43. Association for Computational
  Linguistics, 2020.

\bibitem[Saunders et~al.(2022)Saunders, Sallis, and Byrne]{saunders2022first}
\begingroup\color{blue}
D.~Saunders, R.~Sallis, and B.~Byrne.
\newblock \revised{First the Worst: Finding Better Gender Translations During
  Beam Search}.
\newblock In \emph{\revised{Findings of the Association for Computational
  Linguistics: ACL 2022}}. \revised{Association for Computational Linguistics},
  2022.
\endgroup

\bibitem[Savoldi et~al.(2021)Savoldi, Gaido, Bentivogli, Negri, and
  Turchi]{savoldi2021gender}
B.~Savoldi, M.~Gaido, L.~Bentivogli, M.~Negri, and M.~Turchi.
\newblock Gender bias in machine translation.
\newblock \emph{Transactions of the Association for Computational Linguistics},
  9:\penalty0 845--874, 2021.
\newblock \doi{10.1162/tacl_a_00401}.

\bibitem[Singh(2023)]{singh2023gender}
P.~Singh.
\newblock Gender inflected or bias inflicted: On using grammatical gender cues
  for bias evaluation in machine translation.
\newblock In \emph{Proceedings of the 13th International Joint Conference on
  Natural Language Processing and the 3rd Conference of the Asia-Pacific
  Chapter of the ACL: Student Research Workshop}, pages 17--23, 2023.
\newblock \doi{10.18653/v1/2023.ijcnlp-srw.3}.

\bibitem[Singh et~al.(2024)Singh, Patidar, and Vig]{singh2024translating}
P.~Singh, M.~Patidar, and L.~Vig.
\newblock Translating across cultures: {LLM}s for intralingual cultural
  adaptation.
\newblock In \emph{Proceedings of the 28th Conference on Computational Natural
  Language Learning}, pages 400--418, Miami, FL, USA, 2024. Association for
  Computational Linguistics.
\newblock \doi{10.18653/v1/2024.conll-1.30}.
\newblock URL \url{https://aclanthology.org/2024.conll-1.30/}.

\bibitem[Stafanovi{\v{c}}s et~al.(2020)Stafanovi{\v{c}}s, Bergmanis, and
  Pinnis]{stafanovics2020mitigating}
A.~Stafanovi{\v{c}}s, T.~Bergmanis, and M.~Pinnis.
\newblock Mitigating gender bias in machine translation with target gender
  annotations.
\newblock In \emph{Proceedings of the Fifth Conference on Machine Translation},
  pages 629--638, Online, 2020. Association for Computational Linguistics.
\newblock \doi{10.18653/v1/2020.wmt-1.73}.
\newblock URL \url{https://aclanthology.org/2020.wmt-1.73/}.

\bibitem[Stanovsky et~al.(2019)Stanovsky, Smith, and
  Zettlemoyer]{stanovsky2019evaluating}
G.~Stanovsky, N.~A. Smith, and L.~Zettlemoyer.
\newblock Evaluating gender bias in machine translation.
\newblock In \emph{Proceedings of the 57th Annual Meeting of the Association
  for Computational Linguistics}, pages 1679--1684. Association for
  Computational Linguistics, 2019.
\newblock \doi{10.18653/v1/P19-1164}.

\bibitem[Tiedemann and Thottingal(2020)]{tiedemann2020opusmt}
J.~Tiedemann and S.~Thottingal.
\newblock {OPUS-MT}: Building open translation services for the world.
\newblock In \emph{Proceedings of the 22nd Annual Conference of the European
  Association for Machine Translation}, pages 479--480, 2020.

\bibitem[Vanmassenhove et~al.(2018)Vanmassenhove, Hardmeier, and
  Way]{vanmassenhove2018getting}
\begingroup\color{blue}
E.~Vanmassenhove, C.~Hardmeier, and A.~Way.
\newblock \revised{Getting Gender Right in Neural Machine Translation}.
\newblock In \emph{\revised{Proceedings of the 2018 Conference on Empirical
  Methods in Natural Language Processing}}. \revised{Association for
  Computational Linguistics}, 2018.
\endgroup

\bibitem[Yao et~al.(2024)Yao, Jiang, Bobinac, Yang, and
  Hu]{yao2024benchmarking}
B.~Yao, M.~Jiang, T.~Bobinac, D.~Yang, and J.~Hu.
\newblock Benchmarking machine translation with cultural awareness.
\newblock In \emph{Findings of the Association for Computational Linguistics:
  EMNLP 2024}, pages 13078--13096, Miami, Florida, USA, 2024. Association for
  Computational Linguistics.
\newblock \doi{10.18653/v1/2024.findings-emnlp.765}.
\newblock URL \url{https://aclanthology.org/2024.findings-emnlp.765/}.

\bibitem[Zhao et~al.(2018)Zhao, Wang, Yatskar, Ordonez, and
  Chang]{zhao2018wino}
J.~Zhao, T.~Wang, M.~Yatskar, V.~Ordonez, and K.-W. Chang.
\newblock Gender bias in coreference resolution: Evaluation and debiasing
  methods.
\newblock In \emph{Proceedings of the 2018 Conference of the North American
  Chapter of the Association for Computational Linguistics: Human Language
  Technologies}, pages 15--20. Association for Computational Linguistics, 2018.
\newblock \doi{10.18653/v1/N18-2003}.

\end{thebibliography}
\bibliographystyle{icml2026}
\endgroup

\clearpage
\appendix
\onecolumn

\section{Dataset and Target Subset}
\label{app:dataset}

\begin{table}[H]
\centering
\footnotesize
\renewcommand{\arraystretch}{0.96}
\setlength{\tabcolsep}{4.5pt}
\begin{tabular}{lrll>{\raggedright\arraybackslash}p{0.28\textwidth}}
\toprule
Category & Count & Target? & Source cue & Example type \\
\midrule
explicit\_gender    & 7,500 & Yes & he/she/man/woman & Direct pronoun or gender word \\
late\_binding       & 2,250 & Yes & delayed pronoun & Multi-clause cue \\
winograd\_coref     & 6,000 & Yes & coreference pronoun & WinoBias-style coref \\
name\_profession    & 6,750 & No & Indian name & Name-based cue \\
neutral\_profession & 7,500 & No & none & Ambiguous/gender-neutral \\
counter\_stereotype & 960 & No & varies & Stereotype-violating \\
coreference         & 550 & No & pronoun & Non-Winograd coref \\
multi\_sentence     & 2,340 & No & varies & Multi-sentence context \\
social\_role        & 1,350 & No & role term & Social-role cue \\
temporal\_aspect    & 945 & No & varies & Temporal framing \\
minimal\_context    & 540 & No & none & Minimal context \\
name\_only          & 660 & No & Indian name & Name only \\
\midrule
\textbf{Total}      & \textbf{37,345} & & & \\
\bottomrule
\end{tabular}
\caption{Benchmark category breakdown. Target categories are those used for the preservation-focused mitigation experiments.}
\label{tab:app_category_counts}
\end{table}

\paragraph{Target subset.}
The target subset contains 15,750 examples from \texttt{explicit\_gender},
\texttt{late\_binding}, and \texttt{winograd\_coref} for which the English
source contains a recoverable male or female cue. All three target categories
are balanced 50/50 male/female by construction. We exclude ambiguous or
neutral examples because the cultural-fidelity objective is preservation of
source-explicit gender, not inference of hidden gender.

\begin{table}[H]
\centering
\footnotesize
\renewcommand{\arraystretch}{0.98}
\begin{tabular}{>{\raggedright\arraybackslash}p{0.18\textwidth}>{\raggedright\arraybackslash}p{0.55\textwidth}>{\raggedright\arraybackslash}p{0.14\textwidth}}
\toprule
Category & English source & Gold cue \\
\midrule
explicit\_gender & She started working as a social worker in 2020. & female \\
late\_binding & The engineer completed the project. Later, she received an award. & female \\
winograd\_coref & The engineer met with the nurse because he needed advice. & male \\
\bottomrule
\end{tabular}
\caption{Representative target-subset examples.}
\label{tab:app_examples}
\end{table}

\section{Classifier and Reranking Details}
\label{app:method_details}

\subsection{Automatic Preservation Classifier}
The classifier assigns each Hindi output to one of three preservation states
relative to the English source: \textit{preserved}, \textit{neutralized}, or
\textit{wrong-gender}. It first extracts the expected source-side cue from
English. It then checks the Hindi output for explicit lexical gender markers,
gendered names, gendered verb/adjective morphology, ergative constructions
with \texthindi{ने}, and honorific/plural constructions such as
\texthindi{उन्होंने}. If no reliable rule fires, the example is passed to an
LLM fallback. The classifier is used only as a large-scale diagnostic
instrument; the preservation--fluency frontier is validated through human
judgment.

\begin{algorithm}[H]
\caption{Hindi gender-preservation classifier}
\footnotesize
\begin{algorithmic}[1]
\STATE Extract expected source gender $g$ from the English sentence.
\STATE Detect explicit Hindi lexical gender markers.
\STATE Detect gendered names and gendered kinship/profession terms.
\STATE Detect singular gendered verb/adjective morphology.
\STATE Detect ergative or honorific constructions that remove recoverable subject gender.
\IF{a Hindi cue clearly matches $g$}
    \RETURN preserved
\ELSIF{a Hindi cue clearly conflicts with $g$}
    \RETURN wrong-gender
\ELSIF{the output is fluent but no gender cue remains recoverable}
    \RETURN neutralized
\ELSE
    \RETURN LLM fallback judgment
\ENDIF
\end{algorithmic}
\end{algorithm}

\subsection{Reranking Procedure}
For each source sentence in the target categories, we generate $k=5$ Hindi
candidates using GPT-4o-mini. In the Sarvam setting, the candidate pool
contains the original Sarvam output plus GPT-4o-mini candidates; results for
Sarvam should therefore be interpreted as GPT-assisted repair of Sarvam
translations rather than pure Sarvam generation.

\paragraph{Source-Aware Reranker (SAR).}
SAR selects the candidate with the highest weighted score:
\[
S(y) = \lambda_q\, Q(y) + \lambda_g\, G(x,y) - \lambda_e\, E(y),
\]
where $Q(y)$ estimates translation quality using length ratio, repetition,
and ASCII-density heuristics; $G(x,y)$ rewards preservation of the
source-side gender cue detected by the classifier; and $E(y)$ penalizes
gender-neutralizing ergative or honorific constructions. In the full SAR
configuration, $\lambda_q=0.35$, $\lambda_g=1.0$, and $\lambda_e=0.35$.

\paragraph{Phenomenon-Aware Reranker (PAR).}
PAR generates candidates using phenomenon-specific prompts. It first
classifies the source sentence as \texttt{explicit\_gender},
\texttt{late\_binding}, or \texttt{winograd\_coref} from source-side cues,
then applies a tailored prompt that instructs the model to keep the gender cue
recoverable in Hindi. Unlike SAR, PAR permits minimal lexical gender markers
such as \texthindi{पुरुष} or \texthindi{महिला} when morphological agreement
alone would erase the cue. Selection uses a token-matching score over
predefined male/female Hindi token sets, with a stricter threshold for
multi-clause targets.

\paragraph{Prompt templates.}
\begin{quote}
\small
\textbf{Explicit gender:} Translate the following English sentence into natural
Hindi. The English source explicitly marks the relevant person as
[male/female]. Preserve this gender cue so that a Hindi reader can recover it.
Output only the Hindi translation.

\textbf{Coreference:} Translate the following English sentence into natural
Hindi. Resolve the pronoun/coreference relation in the English source before
translating. Preserve the gender of the referred person in Hindi if the English
source explicitly provides it. Output only the Hindi translation.
\end{quote}

\begin{table}[H]
\caption{Compact SAR ablation on the target subset. Quality-only reranking
stays near Baseline. Gender preservation drives the main gain.}
\label{tab:original_ablation}
\centering
\small
\setlength{\tabcolsep}{6pt}
\begin{tabular}{lrr}
\toprule
Config & GPT target acc. & Sarvam target acc. \\
\midrule
Baseline & 11.07 & 15.99 \\
Prompted L3 & n/a & 14.63 \\
Quality-only SAR & 11.39 & 15.66 \\
Gender-only SAR & 29.24 & 31.68 \\
SAR without ergative penalty & 29.24 & 31.57 \\
SAR & 30.17 & 32.47 \\
\bottomrule
\end{tabular}
\end{table}

\section{Human Evaluation Protocol}
\label{app:human_protocol}

Annotators saw the English source sentence and two anonymized Hindi
translations in randomized order. They did not know which output came from
the Baseline or reranked system. For each Hindi output, annotators answered:
\begin{center}
\begin{tabular}{@{}r@{\quad}p{0.86\linewidth}@{}}
1. & Does the Hindi translation preserve the gender cue explicitly expressed in the English source? \\
2. & How fluent and natural is the Hindi translation on a 1--5 scale? \\
3. & Which translation is better overall: A, B, or tie?
\end{tabular}
\end{center}
A translation was marked as preserving gender only if a Hindi reader could
recover the male or female cue expressed in the English source. A fluent
translation was still marked as non-preserving if the explicit source-side
gender cue became unrecoverable.

\begin{table}[H]
\centering
\small
\begin{tabular}{cp{0.72\linewidth}}
\toprule
Score & Meaning \\
\midrule
1 & Ungrammatical or very unnatural Hindi \\
2 & Understandable but clearly awkward \\
3 & Acceptable but noticeably unnatural \\
4 & Mostly natural Hindi \\
5 & Fully natural/native-quality Hindi \\
\bottomrule
\end{tabular}
\caption{Human fluency scale.}
\label{tab:app_fluency_scale}
\end{table}

\begin{table}[H]
\caption{Per-category human results for PAR, pooled across both raters. The
strongest fluency cost appears in late-binding constructions.}
\label{tab:human_category}
\centering
\small
\setlength{\tabcolsep}{5pt}
\begin{tabular}{lrrrr}
\toprule
Category & Pres. PAR & Pres. B & Flu. PAR & Flu. B \\
\midrule
Explicit gender & 80.0 & 15.0 & 4.06 & 4.15 \\
Late binding & 77.0 & 6.0 & 2.67 & 4.57 \\
Winograd coref & 87.0 & 10.0 & 3.37 & 4.35 \\
\bottomrule
\end{tabular}
\end{table}

The human study is not intended as a deployment-level estimate of translation
quality. Its purpose is to validate that the automatic preservation signal
corresponds to Hindi-fluent reader judgments and that the fluency cost
introduced by lexicalization is perceptible to humans.

\section{Qualitative Examples and Failure Modes}
\label{app:qualitative}

Table~\ref{tab:app_qualitative} illustrates the four outcome types on the
preservation--fluency frontier. The first two rows show successful interventions
(SAR avoidance and PAR bypass), while the last two rows show cases where PAR
either harms fluency or adds a redundant marker.

\begin{table}[H]
\centering
\footnotesize
\setlength{\tabcolsep}{4pt}
\begin{tabular}{>{\raggedright\arraybackslash}p{0.12\textwidth}%
                >{\raggedright\arraybackslash}p{0.18\textwidth}%
                >{\raggedright\arraybackslash}p{0.23\textwidth}%
                >{\raggedright\arraybackslash}p{0.23\textwidth}%
                >{\raggedright\arraybackslash}p{0.14\textwidth}}
\toprule
Type & English & Baseline Hindi & Reranked Hindi & Notes \\
\midrule
Successful avoidance (SAR) &
She started working as a social worker in 2020. &
\texthindi{उसने 2020 में एक सामाजिक कार्यकर्ता के रूप में काम करना शुरू किया।} &
\texthindi{वह 2020 में सामाजिक कार्यकर्ता के रूप में कार्यरत हुई।} &
Feminine \texthindi{हुई} preserves gender; no explicit marker needed. \\
\addlinespace
Successful bypass (PAR) &
She is a skilled babysitter during the planning phase. &
\texthindi{वह योजना चरण के दौरान एक कुशल बेबीसिटर है।} &
\texthindi{वह योजना चरण के दौरान एक कुशल महिला बेबीसिटर है।} &
Copula \texthindi{है} is genderless; \texthindi{महिला} acts as a lexical bypass. \\
\addlinespace
Fluency cost (PAR) &
The clerk completed the project. Later, he received an award. &
\texthindi{क्लर्क ने परियोजना पूरी की। बाद में, उसे एक पुरस्कार मिला।} &
\texthindi{क्लर्क ने प्रोजेक्ट पूरा किया। बाद में, वह पुरुष ने पुरस्कार प्राप्त किया।} &
\texthindi{वह पुरुष ने} is unnatural as an anaphor; PAR preserves gender but harms fluency. \\
\addlinespace
Baseline better (PAR) &
She recently became an engineer. &
\texthindi{वह हाल ही में इंजीनियर बनी।} &
\texthindi{वह हाल ही में एक महिला इंजीनियर बनी।} &
\texthindi{बनी} is already feminine; adding \texthindi{महिला} is redundant. \\
\bottomrule
\end{tabular}
\caption{Representative examples illustrating the preservation--fluency
frontier. The successful avoidance example illustrates SAR, while the bypass
and failure examples illustrate PAR. The fluency-cost row is an intentionally
selected failure case illustrating why PAR is not a dominant solution.}
\label{tab:app_qualitative}
\end{table}

\end{document}